\documentclass[letterpaper]{article} 
\usepackage[preprint]{aaai2027}  
\usepackage[hyphens]{url}  
\usepackage{graphicx} 
\usepackage{natbib}  
\usepackage{caption} 
\usepackage{algorithm}
\usepackage{algorithmic}

\usepackage{newfloat}
\usepackage{listings}
\DeclareCaptionStyle{ruled}{labelfont=normalfont,labelsep=colon,strut=off} 
\floatstyle{ruled}
\newfloat{listing}{tb}{lst}{}
\floatname{listing}{Listing}

\usepackage{booktabs}

\usepackage{amsmath}
\usepackage{tcolorbox}
\usepackage{makecell}
\usepackage{multirow}
\usepackage{colortbl}
\usepackage{pgfplots}
\usepackage{subcaption}
\pgfplotsset{compat=1.17}
\usepgfplotslibrary{groupplots} 

\title{Test-Time Self-Evolving GUI Visual Grounding via Reflection-Guided On-Policy Self-Distillation}
\author{
    Shiyu Xuan\textsuperscript{\rm 1}, Zechao Li\textsuperscript{\rm 1}\\
}
\affiliations{
    \textsuperscript{\rm 1}School of Computer Science and Engineering, Nanjing University of Science and Technology, Nanjing, China, 210094\\
    \{shiyu\_xuan, zechao.li\}@njust.edu.cn
}

\begin{document}

\maketitle

\begin{abstract}
GUI Visual Grounding is a fundamental capability for GUI agents. Existing models typically freeze their parameters after deployment, limiting their ability to adapt to unseen interfaces. Although recent methods attempt to adapt models via test-time reinforcement learning, they cannot reflect upon failed exploration. To overcome this, we propose a Test-Time Self-Evolving framework that enables models to improve after deployment without human-annotated ground truth. It constructs a closed-loop of Exploration, Evaluation, Reflection, and Internalization. Specifically, the agent first explores unseen interfaces by predicting grounding coordinates for given instructions. To evaluate these explorations, we introduce an MLLM-based Reflector to assess the generated results and provide the corresponding reasoning reflections.
To internalize reflection knowledge into the model weights, we propose Reflection-Guided On-Policy Self-Distillation, which translates high-level reasoning into dense token-level supervision via a conditioned self-teacher. Furthermore, we design a Contrastive Calibration method to prevent incorrect auto-regressive prefixes from corrupting the supervisory signals during failed explorations. Extensive experiments across six benchmarks demonstrate our framework's effectiveness, achieving an average accuracy improvement of 7.4\% over the base model. To the best of our knowledge, this is the first work to successfully exploit on-policy self-distillation for test-time adaptation in GUI visual grounding. By filling the gap in post-deployment adaptation, our framework completes the self-evolving capability of GUI agents. The code will be released.

\end{abstract}


\section{Introduction}
GUI agents capable of navigating and interacting with digital environments have recently emerged as a promising direction for human-computer interaction~\cite{osworldg}. As a fundamental capability of GUI agents, GUI Visual Grounding aims to localize UI elements according to natural-language instructions. Recent advances have substantially improved grounding performance through large-scale supervised fine-tuning~\cite{uitars,cheng2024seeclick}, specialized action heads~\cite{Gui-actor,lin2025showui}, and reinforcement learning~\cite{gui-r1,se-gui}.

\begin{figure}[t!]
\centering
\includegraphics[width=1.0\columnwidth]{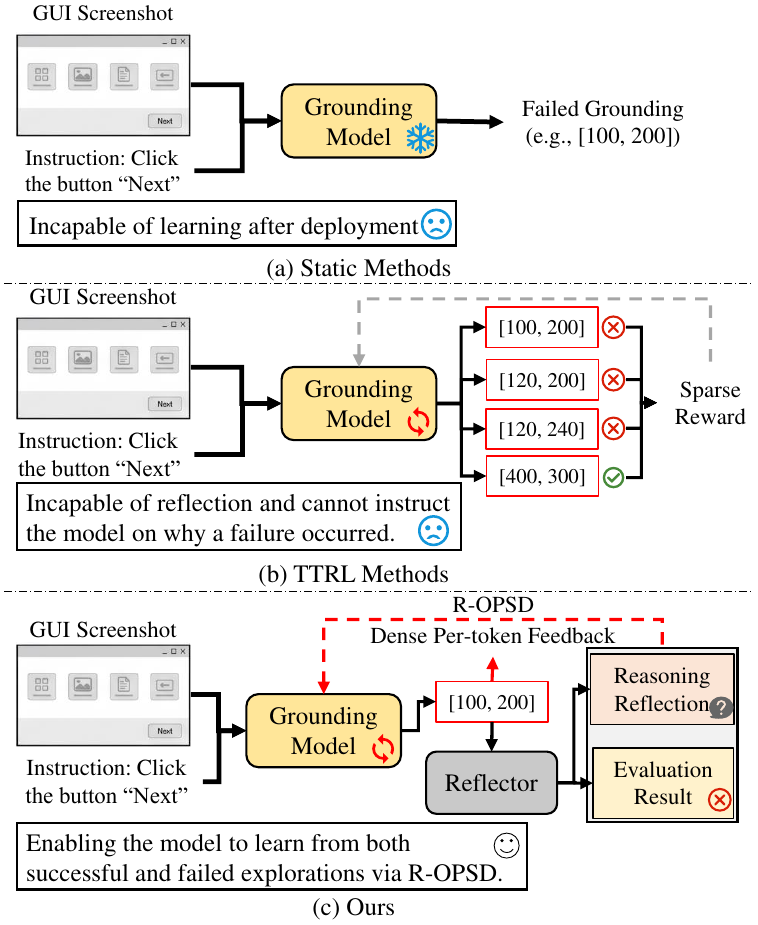}
\caption{(a) Static methods operate on frozen parameters after deployment and cannot learn from exploration. (b) TTRL Methods adapt the model using sparse scalar rewards, but lack a mechanism to reason about failures and internalize textual reflections. (c) Test-Time Self-Evolving introduces a Reflector to provide detailed textual reflections, which are translated into dense token-level feedback via R-OPSD, enabling the model to self-evolve.}
\label{fig:motivation}
\end{figure}

Despite these advances, existing GUI grounding models remain static after deployment, with model parameters frozen once training is completed. Consequently, when encountering unseen applications or interface layouts, their performance cannot improve through interaction. A recent test-time adaptation method~\cite{du2026test} attempts to alleviate this limitation by optimizing the model online with reinforcement learning. This method relies on sparse scalar rewards that merely indicate whether an exploration succeeds. Such feedback provides little information about \emph{why} a prediction fails or how the model should correct it.

A natural solution is to train the model with the reflection on its own explorations. Rather than merely assigning a success or failure signal, reflection explains why a prediction is incorrect by identifying the intended UI element, examining the predicted location, and diagnosing the failure. Such reasoning provides richer supervisory information than scalar rewards. However, existing reinforcement learning algorithms cannot directly leverage these textual reflections based on scalar rewards.

To overcome this, we propose a Test-Time Self-Evolving framework. It constructs a closed-loop that equips the agent with four core capabilities: \textbf{Exploration} in unknown interfaces, \textbf{Evaluation} of generated coordinates, \textbf{Reflection} upon evaluation results, and \textbf{Internalization} of these reflections into its parameters. Specifically, upon deployment, the agent first explores the interface with instructions by predicting grounding coordinates. To evaluate this exploration, we introduce an MLLM-based Reflector that estimates whether the prediction is consistent with the instruction while simultaneously generating a reflection explaining its evaluation.

The remaining challenge is how to internalize the reflection into the model parameters. Unlike scalar rewards, textual reflections cannot be directly exploited by policy optimization algorithms. To bridge this gap, we propose Reflection-Guided On-Policy Self-Distillation (R-OPSD). Instead of learning from scalar rewards, R-OPSD conditions a self-teacher on the Reflector's evaluation results and reflections, converting high-level reasoning into dense token-level supervision over the generated coordinate tokens.

However, this dense feedback encounters a bottleneck in the auto-regressive generation of coordinates. During failed explorations, the model predicts incorrect tokens. If the teacher model conditions its subsequent supervision on these prefixes, the supervisory signals gradually become unreliable. To mitigate this issue, we propose a Contrastive Calibration Method. 
It treats a failed exploration as successful by using an inverse-prompted student.
The trajectory is sampled from the model's own policy, conditioning it on a success prompt amplifies its confidence in this specific sequence. Consequently, at the initial error step, the inverse student assigns a high probability to the incorrect token, while the reflection-guided teacher yields a lower probability. This discrepancy produces a negative advantage that suppresses the initial error. As generation continues, the corrupted prefix dominates both models, aligning their output distributions. This decays the advantage to almost zero, preventing corrupted feedback from being internalized.

Extensive experiments across ScreenSpot and ScreenSpot-v2~\cite{cheng2024seeclick}, ScreenSpot-Pro~\cite{screenspotpro}, MMBench-GUI~\cite{mmbench}, OSWorld-G and OSWorld-G-Refine~\cite{osworldg} demonstrate the effectiveness of our framework. By enabling self-evolving without ground truth annotations, our method achieves an average accuracy improvement of 7.4\% over the base model. Furthermore, compared to the recent test-time reinforcement learning method GUI-RCPO, our method achieves an average performance gain up to 7.7\%.

In summary, our main contributions are three-fold: (a) We introduce a Test-Time Self-Evolving framework, which empowers grounding agents with exploration, evaluation, reflection, and parameter internalization in unseen environments. (b) We propose Reflection-Guided On-Policy Self-Distillation, which converts textual reflections into dense token-level supervision, together with a Contrastive Calibration method that alleviates corrupted supervision caused by incorrect auto-regressive prefixes.
(c) Extensive experiments and ablations validate our framework. This is the first work to applying OPSD in test-time scenarios of GUI grounding, establishing a promising direction for GUI grounding.

\section{Related Works}
\noindent\textbf{GUI Visual Grounding.}
GUI visual grounding locates UI elements based on natural language instructions. Recent advancements span three paradigms. \textbf{Supervised fine-tuning methods} establish baseline alignment using large-scale datasets~\cite{uitars, cheng2024seeclick, gou2024navigating}. To enhance visual perception, models selectively extract UI-related visual tokens~\cite{lin2025showui, ouyang2026focusui} or employ attention-based action heads to bypass direct coordinate generation~\cite{Gui-actor}. \textbf{Reinforcement learning methods} further enhance the performance. Recent works adapt GRPO~\cite{shao2024deepseekmath} with rule-based coordinate rewards~\cite{gui-r1, ui-r1}, or design complex reward functions incorporating attention maps, box sizes, and bounding-box distributions~\cite{se-gui, gui-g1, GUI-G$^2$}. \textbf{Test-time scaling methods} allocate additional inference compute to enhance performance. RegionFocus~\cite{luo2025visual} dynamically zooms into the image for coarse-to-fine UI grounding, a process similarly guided by attention maps in ZoomUI~\cite{zoomui}. Additionally, GUI-RC~\cite{du2026test} aggregates multiple predictions via region consistency, further refining the model with test-time reinforcement learning~\cite{zuo2026ttrl}.

\noindent\textbf{On-policy Distillation.}
To internalize reasoning capabilities into model weights, on-policy distillation (OPD) trains the student model with token-level feedback on its own generation trajectories. Typically, this feedback is derived either from a stronger teacher model~\cite{gu2024minillm,agarwal2024policy} or a model conditioned on privileged information~\cite{zhang2026opsdl,zhao2026self}.
However, incorrect prefixes generated by the student can degrade the quality of the feedback. To address this issue, ESR~\cite{ziheng2026less} proposes an early stopping rollout mechanism, while EOPD~\cite{jin2026entropy} assesses the reliability of the teacher model using token entropy. Alternatively, RLSD~\cite{yang2026self} and RLCSD~\cite{pan2026rlcsd} leverage token-level feedback to only determine update magnitudes. The potential of OPD has also been explored across various vision-language tasks, such as Video Temporal Grounding~\cite{li2026video}, Visual Question Answering~\cite{yuan2026vision}, and GUI Visual Grounding~\cite{zhang2026learn}.

Unlike test-time reinforcement learning methods~\cite{du2026test, zuo2026ttrl} that rely on sparse scalar rewards and fail to leverage reflection reasoning, our framework empowers the model to learn from rich reflections via R-OPSD. Furthermore, Contrastive Calibration effectively migrates the corrupted supervision issue caused by incorrect prefixes through suppressing initial errors and decaying the policy gradients of drifted tokens, ensuring that the model can safely internalize knowledge from failed explorations.

\section{Methodology}
\subsection{Overview}
\begin{figure*}[t] 
    \centering
    \includegraphics[width=0.98\textwidth]{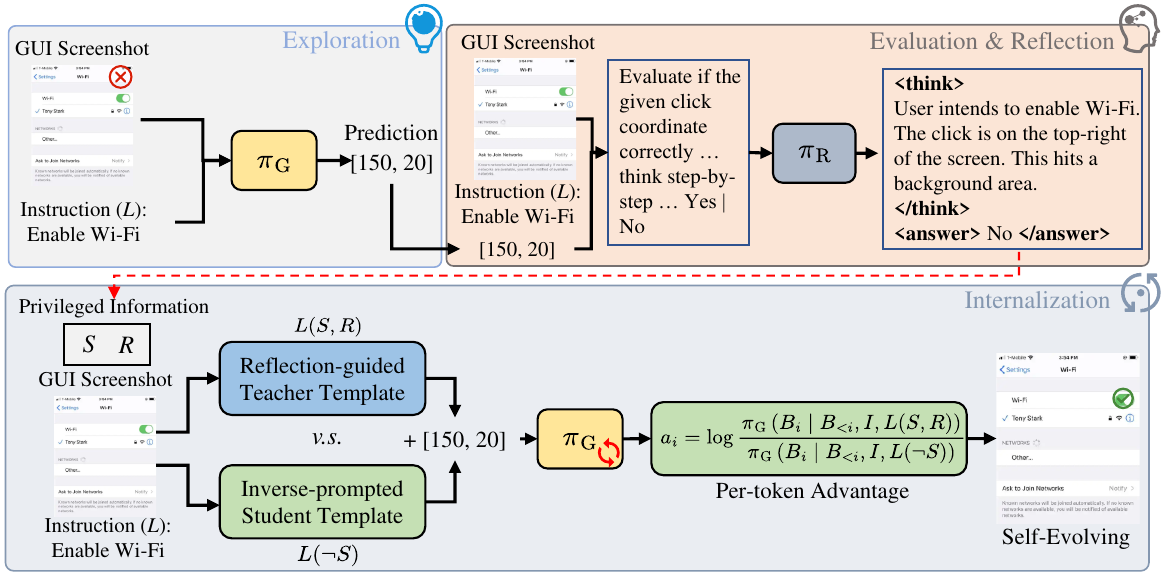}
    \caption{\textbf{Test-Time Self-Evolving Framework} consists of four core stages: Exploration, Evaluation, Reflection, and Internalization. During test-time adaptation, $\pi_{\text{G}}$ explores the environment to predict coordinates. The Reflector $\pi_{\text{R}}$ then assesses the prediction to generate a binary score $S$ and a step-by-step reasoning $R$. The reflection knowledge is internalized via Reflection-Guided On-Policy Self-Distillation to update the grounding policy. To optimize memory efficiency, both $\pi_{\text{G}}$ and $\pi_{\text{R}}$ share a base model and we alternate between the roles of $\pi_{\text{G}}$ and $\pi_{\text{R}}$ by switching the active LoRA adapters.}
    \label{fig:framework}
\end{figure*}

Given a screenshot $I$ and a language instruction $L$, the GUI grounding model $\pi_{\text{G}}$ is required to locate the corresponding UI element $B = \pi_{\text{G}}(I, L)$, where $B$ denotes the coordinates of the bounding box $[x_1, y_1, x_2, y_2]$ or the center point $(x, y)$ of the target element. Existing GUI grounding models typically remain static after deployment, with model parameters frozen once training is completed. Consequently, when deployed on unseen platforms or applications, the model cannot improve from its own interaction experience.

To enable post-deployment adaptation without human annotations, we propose a Test-Time Self-Evolving framework that constructs a closed-loop consisting of four stages: \textbf{Exploration} in unknown interfaces, \textbf{Evaluation} of its generated coordinates, \textbf{Reflection} upon its evaluation result, and \textbf{Internalization} of these reflections into its parameters. Specifically, this framework executes an episodic loop:
\begin{itemize}
    \item \textbf{Exploration:} The GUI grounding model $\pi_{\text{G}}$ explores the unseen interface, \emph{i.e.}, given the UI screenshot $I$ and instruction $L$, $B=\pi_{\text{G}}(I,L)$.
    \item \textbf{Evaluation \& Reflection:} To assess the exploration in the absence of ground truth, we introduce a Reflector $\pi_{\text{R}}$. It takes the screenshot, instruction, and predicted coordinates as input to output an evaluation score $S$ and a detailed reasoning process $R$,
    \begin{equation}
        \small
        S, R = \pi_{\text{R}}(I, L, B),
    \end{equation}
    where $S \in \{0, 1\}$ indicates whether the current exploration is successful, and $R$ represents the reasoning behind this evaluation. Subsequently, $S$ and $R$ are utilized to guide the internalization.
    \item \textbf{Internalization:} The reflection knowledge is then internalized into $\pi_{\text{G}}$ through Reflection-Guided On-Policy Self-Distillation. Specifically, the evaluation score and reasoning process are leveraged as \emph{privileged information} to construct a self-teacher that provides per-token feedback on the generated coordinates. Furthermore, to alleviate failed explorations from corrupting the feedback, we propose a \textbf{Contrastive Calibration Method}.
\end{itemize}

In the following parts, we will first introduce the Evaluation \& Reflection stages, and further propose the Internalization stage. The overall framework is illustrated in Fig.\ref{fig:framework}.

\subsection{Evaluation and Reflection with Reflector}\label{sec:reflector}
During test-time adaptation, ground-truth annotations are unavailable. Therefore, the model requires an alternative mechanism to estimate whether its exploration is successful.
To this end, we introduce an MLLM-based Reflector $\pi_{\text{R}}$. Its role is to evaluate the alignment between the predicted coordinates $B$ and the language instruction $L$ within the context of the visual screenshot $I$. Besides producing an estimated evaluation result, it also generates a reflection explaining the reasoning behind its judgment. This reflection serves as privileged information for the subsequent internalization stage.

\noindent\textbf{Inference.} To achieve this goal, we design a prompt template that forces the Reflector to engage in step-by-step reasoning before outputting a final result:
\begin{tcolorbox}[
    colback=red!5!white,
    colframe=red!80!black,
    boxrule=1pt,
    arc=1mm,
    left=2mm, right=2mm, top=1mm, bottom=1mm,
    fonttitle=\bfseries\small,
    fontupper=\footnotesize,
    title=\textbf{Reflector Prompt Template}
]
Evaluate if the given click coordinate correctly executes the user instruction on the provided UI image. Answer with Yes or No.\\
Instruction: \{$L$\}\\
Click Coordinate: \{$B$\}\\
Before providing your final answer, you must think step-by-step. Address the following in your reasoning:\\
1. Intent Analysis: What specific UI element (text, icon, button) does the instruction intend to interact with?\\
2. Coordinate Grounding: Look at the exact location of the Click Coordinate [x,y] on the UI. What specific element or visual feature is located at that coordinate?\\
3. Verification: Does the element at the given coordinate match the intended element from the instruction? Is it the correct target, or did it hit the background, a different element, or an empty space?\\
4. Conclusion: Explicitly state whether the verification passed (Yes) or failed (No).\\[1pt]
Output the thinking process in <think> </think> and final answer in <answer> </answer> tags.\\
You must strictly format your output as follows:\\
<think> Your step-by-step reasoning and conclusion here </think>\\
<answer>Yes|No</answer>
\end{tcolorbox}

Once receiving the generated response from $\pi_{\text{R}}$, we parse the output according to the specified tags to obtain $R$ and $S$.
Specifically, the text extracted from the \texttt{<think>...</think>} tags serves as the reflection $R$,
The discrete binary label extracted from the \texttt{<answer>...</answer>} tags serves as $S$.

\noindent\textbf{Training.}
To improve the reliability of the Reflector, we optimize it using GRPO~\cite{shao2024deepseekmath} on an offline collected dataset.
Each sample in this dataset comprises a UI screenshot $I$, a language instruction $L$, a sampled click coordinate $B$, and a ground-truth label $S^* \in \{0, 1\}$ indicating whether the coordinate successfully executes the instruction. The construction details of the dataset are shown in the \textbf{Appendix}.
We employ the format and binary rewards during training, \emph{i.e.}, $r = r_{\text{format}} + r_{\text{binary}}$:
\begin{itemize}
    \item $r_{\text{format}}$ ensures the model strictly adheres to the requested output template.
    \item $r_{\text{binary}}$ supervises discrimination capability:
    \begin{equation}
    \small
        r_{\text{binary}} =
        \begin{cases} 
        1, & \text{if } S = S^* \text{ and format is valid} \\
        0, & \text{if } S \neq S^* \text{ or format is invalid}
        \end{cases}
    \end{equation}
\end{itemize}
After training, the Reflector remains frozen throughout test-time adaptation.

\subsection{Internalization with Reflection-Guided On-Policy Self-Distillation}
After obtaining the evaluation result $S$ and reflection $R$, the final stage is to internalize this knowledge into the grounding model $\pi_{\text{G}}$.
The remaining challenge is how to transform textual reflection into effective optimization signals.
Unlike reinforcement learning, which is based on scalar rewards, reflection consists of free-form natural language and therefore cannot be directly exploited by standard policy optimization algorithms.
To bridge this gap, we propose Reflection-Guided On-Policy Self-Distillation (R-OPSD), which converts reflection into dense token-level supervision.

\noindent\textbf{R-OPSD.}
Given a rollout $B \sim \pi_{\text{G}}(\cdot \mid I, L)$ generated by the policy, OPD trains the policy $\pi_{\text{G}}$ on its own rollout with the token-level supervisory signals provided by a teacher model $\pi_{\text{T}}$.
Following the standard policy optimization formulation, the token-level advantage is defined as:
\begin{equation}
    \small
    a_i = \log \frac{\pi_{\text{T}}(B_i \mid B_{<i}, I, L)}{\pi_{\text{G}}(B_i \mid B_{<i}, I, L)}.
\end{equation}
The optimization objective maximizes the expected return, defined as the sum of token-level advantages:
\begin{equation}
    \small
    \mathcal{L}_{\text{OPD}} = - \frac{1}{T} \sum_{i=1}^{T} \text{sg}(a_i) \log \pi_{\text{G}}(B_i \mid B_{<i}, I, L),
    \label{eq:OPD}
\end{equation}
where $\text{sg}(\cdot)$ denotes the stop-gradient operator.

To internalize the knowledge from the Reflector, we construct a self-teacher by utilizing the $\pi_{\text{G}}$ itself conditioned on the evaluation result $S$ and reflection $R$ as the privileged information. Conditioning on this additional information enables the teacher to produce a more informative conditional distribution over the generated coordinate sequence.
Specifically, the input prompt $L(S, R)$ is formulated as follows,
\begin{tcolorbox}[
    colback=red!5!white,
    colframe=red!80!black,
    boxrule=1pt,
    arc=1mm,
    left=2mm, right=2mm, top=1mm, bottom=1mm,
    fonttitle=\bfseries\small,
    fontupper=\footnotesize,
    title=\textbf{Reflection-guided Teacher Prompt Template}
]
\textbf{If Success ($S=1$)} \\
The previous prediction $\{B\}$ has been verified to correctly fulfill the instruction. Outline the position corresponding to the instruction: $\{L\}$. The output should be only [x1,y1,x2,y2]. \\ \\
\textbf{If Failure ($S=0$)} \\
Previous wrong prediction: $\{B\}$ \\
Feedback: $\{R\}$ \\ 
Based on the feedback, outline the position corresponding to the instruction: $\{L\}$. The output should be only [x1,y1,x2,y2].
\end{tcolorbox}

The token-level advantage is then computed as:
\begin{equation}
    \small
    a_i = \log \frac{\pi_{\text{G}}\left(B_i \mid B_{<i}, I, L(S, R)\right)}{\pi_{\text{G}}\left(B_i \mid B_{<i}, I, L\right)}.
    \label{eq:OPD-Reflection-advantage}
\end{equation}

\noindent\textbf{Contrastive Calibration Method.}
MLLM-based GUI grounding is formulated as an auto-regressive generation of coordinate tokens. Performing R-OPSD on failed explorations introduces a challenge. If the teacher model conditions its subsequent feedback on these incorrect prefixes, its predicted probabilities become meaningless.

\begin{figure}[t]
    \centering
    \includegraphics[width=0.95\linewidth]{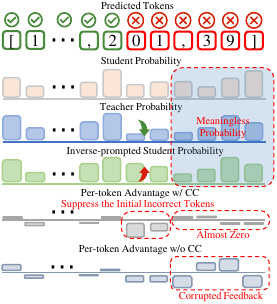}
    \caption{Illustration of Contrastive Calibration Method. During auto-regressive generation, an incorrect prediction, \emph{e.g.,} the initial incorrect token `0`, causes subsequent tokens to drift. Without CC, computing advantages on these drifted tokens results in corrupted feedback. By contrast, CC leverages an inverse-prompted student. This produces a negative penalty to suppress the initial incorrect tokens. As errors accumulate, the conditioned incorrect prefix forces both the teacher and the inverse-prompted student's distributions to align, decaying the advantage to near zero.}
    \label{fig:contrastive_calibration}
\end{figure}

To mitigate this, we propose a \textbf{Contrastive Calibration Method (CC)} to calibrate the learning advantage based on the state of the generated prefix: it suppresses the initial incorrect token and decays the advantage to zero as the tokens progressively drift. To achieve this, we introduce an \emph{inverse-prompted student} to contrast against the teacher.
Specifically, for failed explorations, we construct a contrastive prompt $L(\neg S)$ to misinform the model that the prediction is correct,
\begin{tcolorbox}[
    colback=red!5!white,
    colframe=red!80!black,
    boxrule=1pt,
    arc=1mm,
    left=2mm, right=2mm, top=1mm, bottom=1mm,
    fonttitle=\bfseries\small,
    fontupper=\footnotesize,
    title=\textbf{Inverse-prompted Student Prompt Template}
]
\textbf{If Failure ($S=0$)} \\
\textcolor{red}{Previous correct prediction: $\{B\}$} \\
Based on the feedback, outline the position corresponding to the instruction: $\{L\}$. The output should be only [x1,y1,x2,y2].
\end{tcolorbox}

With $L(\neg S)$, we reformulate the advantage in Eq.\eqref{eq:OPD-Reflection-advantage} into the contrastive calibration advantage:
\begin{equation}
    \small
    a_i = \log \frac{\pi_{\text{G}}\left(B_i \mid B_{<i}, I, L(S, R)\right)}{\pi_{\text{G}}\left(B_i \mid B_{<i}, I, L(\neg S)\right)}.
\end{equation}
Note that CC is exclusively applied to failed explorations, as successful samples ($S=1$) avoid the incorrect prefix issue.

As shown in Fig.~\ref{fig:contrastive_calibration}, CC addresses the incorrect prefix issue across two phases. First, when the prefix $B_{<i}$ is correct, but the current token $B_i$ is incorrect, the inverse-prompted student under $L(\neg S)$ is misled to treat the prediction as successful, assigning a high probability to $B_i$. In contrast, the teacher under $L(S, R)$ leverages the reflection to predict the true target, yielding a lower probability for $B_i$, producing a negative advantage $a_i < 0$, driving the policy gradient to suppress this token. Second, As more incorrect tokens are generated, both models are conditioned on increasingly similar prefixes, leading to similar probabilities $\pi_{\text{G}}(\cdot \mid L(S, R)) \approx \pi_{\text{G}}(\cdot \mid L(\neg S))$, driving the advantage to almost zero ($a_i \to 0$), thereby gradually reducing the magnitude of the token-level advantage.

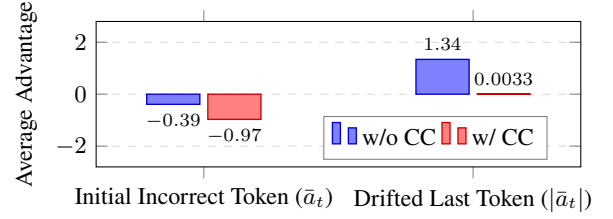
\begin{figure}[t]
\centering
\begin{tikzpicture}
    \begin{axis}[
        ybar=3pt,
        width=0.95\linewidth,
        height=3.5cm,
        enlarge x limits=0.4,
        ylabel={Average Advantage}, 
        symbolic x coords={Initial Token, Last Token},
        xtick=data,
        xticklabels={Initial Incorrect Token ($\bar{a}_t$), Drifted Last Token ($|\bar{a}_t|$)},
        nodes near coords,
        nodes near coords align={vertical},
        every node near coord/.append style={font=\scriptsize,/pgf/number format/fixed,/pgf/number format/precision=5},
        ymin=-2.8, ymax=2.8,
        bar width=20pt,
        legend style={
            at={(0.7,0.35)},
            anchor=north,
            legend columns=-1,
            draw=black!50,
            fill=white,
            font=\small
        },
        ymajorgrids=true,
        grid style={dashed, gray!30},
        tick label style={font=\small, align=center}, 
        label style={font=\small}
    ]

    \addplot[fill=blue!50, draw=blue!80!black, line width=0.5pt] coordinates {
        (Initial Token, -0.39)  
        (Last Token, 1.34)
    };
    \addlegendentry{w/o CC}

    \addplot[fill=red!50, draw=red!80!black, line width=0.5pt] coordinates {
        (Initial Token, -0.97) 
        (Last Token, 0.0033)   
    };
    \addlegendentry{w/ CC}

    \end{axis}
\end{tikzpicture}
\caption{Quantitative analysis of token-level advantage on failed explorations ($S=0$). We report the average advantage ($\bar{a}_t$) at the initial incorrect token, and the average absolute advantage ($|\bar{a}_t|$) at the drifted last tokens.}
\label{fig:advantage_stats}
\end{figure}

\noindent\textbf{Discussion.}
To validate CC, we analyze token-level advantage during failed explorations by calculating the average advantage ($\bar{a}_t$) at the initial incorrect token and the average absolute advantage ($|\bar{a}_t|$) at the drifted last token. As shown in Fig.~\ref{fig:advantage_stats}, without CC, R-OPSD applies a marginal penalty at the initial error and generates corrupted advantages $|\bar{a}_t| = 1.34$ on drifted tokens.
In contrast, CC generates a strong negative penalty at the initial incorrect token. As tokens progressively drift, the update magnitude decays towards zero, preventing the internalization of corrupted supervision.

\noindent\textbf{Direction-based Advantage Clamping.}
Directly applying token-level feedback of R-OPSD can sometimes destabilize training. To mitigate this, the token-level updates should be aligned with the evaluation result, \emph{i.e.}, if a prediction is successful, its probability should be encouraged. If it fails, its probability should be suppressed. Therefore, we apply direction-based clamping to the advantage,
\begin{equation}
    \small
    a_i = \left\{\begin{aligned}
            &\max(0, a_i), & S=1 \\
            &\min(0, a_i), & S=0
        \end{aligned}
    \right..
\end{equation}

In addition, this token-level advantage can be seamlessly integrated with the query-level advantage from GRPO.
Given a sampled group of size $K$, the GRPO reward is defined as $r_{\text{GRPO}} = r_{\text{format}} + r_{\text{correctness}}$, where $r_{\text{correctness}}$ is evaluation $S$ of the predicted coordinates. The query-level advantage $\hat{A}_{\text{GRPO}}[k]$ is computed by standardizing these rewards:
\begin{equation}
    \small
    \hat{A}_{\text{GRPO}}[k] = \frac{r_{\text{GRPO}}[k] - \text{mean}\{r_{\text{GRPO}}[1], \dots, r_{\text{GRPO}}[K]\}}{\text{std}\{r_{\text{GRPO}}[1], \dots, r_{\text{GRPO}}[K]\}}.
\end{equation}
By incorporating the token-level advantage into the GRPO advantage, the final integrated objective is:
\begin{equation}
    \small
    \hat{A}[k] = \hat{A}_{\text{GRPO}}[k] + \lambda a_i,
    \label{eq:OPD-Reflection-advantage-Contrastive}
\end{equation}
where $\lambda$ controls the integration strength.

Although the proposed framework relies on the Reflector to estimate grounding correctness, it does not require perfect evaluation. Instead, R-OPSD only assumes that the Reflector provides sufficiently informative evaluations and reflections to improve the conditional distribution of the teacher. 
As demonstrated in our experiments, this assumption is sufficient to improve grounding performance across multiple GUI benchmarks consistently. We also provide quantitative and qualitative evaluations of the Reflector in the \textbf{Appendix}.

\section{Experiments}

\begin{table*}[t!]
\centering
\small
\setlength{\tabcolsep}{8.5px}
\begin{tabular}{l cccccccc}
\toprule
\textbf{Method} & \makecell{\textbf{Adaptation} \\ \textbf{Datasets}} & \textbf{SS} & \textbf{SSv2} & \textbf{SSP} & \textbf{OSW-G} & \textbf{OSW-GR} & \textbf{MMG} & \textbf{Avg.} \\
\midrule
\multicolumn{9}{l}{\textit{GUI-specific Models}} \\ \midrule
UGround-V1-2B~\cite{gou2024navigating} & - & 77.7 & 78.8 & 22.7 & - & - & - & - \\
ShowUI-2B~\cite{lin2025showui} & - & 75.1 & - & - & - & - & - & - \\
GUI-Actor-3B~\cite{Gui-actor} & - & 89.7 & 91.0 & 42.2 & - & - & - & - \\
FocusUI-3B~\cite{ouyang2026focusui} & - & - & 91.5 & 43.8 & 53.4 &  - & - & - \\ \midrule
InfiGUI-R1-3B~\cite{infigui-r1} & -  & 87.5  & -  & 35.7 & -    & -    & -    & -  \\
SE-GUI-3B~\cite{se-gui} & -    & -  &-  & 35.9 & -    & -    & -    & -    \\
GUI-G$^2$-3B~\cite{GUI-G$^2$} & -     & 90.8    & 91.0 & 37.8 & -    & -    & -    & -  \\
\midrule
\multicolumn{9}{l}{\textit{Test-Time Adaptation}} \\ \midrule
Qwen2.5-VL-3B \cite{qwen2.5-VL}& - & 78.3 & 80.4 & 20.3 & 27.1 & 37.7 & 57.5 & 50.2 \\
+ GUI-RCPO~\cite{du2026test} & SSv2 & 82.3 & 85.4 & 24.8 & 29.3 & 38.4 & 60.2 & 53.4\footnotesize{+3.2} \\
\rowcolor{gray!20} \textbf{+ Ours} & SSv2 & \textbf{85.9} & \textbf{88.8} & \textbf{30.5} & 32.6 & 39.7 & 66.6 & 57.4\footnotesize{+7.2} \\
+ GUI-RCPO~\cite{du2026test} & MMG & 79.2 & 80.6 & 19.8 & 25.7 & 36.1 & 58.2 & 49.9\footnotesize{-0.3} \\
\rowcolor{gray!20} \textbf{+ Ours} & MMG & 84.6 & 87.5 & 30.1 & \textbf{33.7} & \textbf{41.7} & \textbf{68.2} & \textbf{57.6\footnotesize{+7.4}} \\ \midrule
Qwen3-VL-2B \cite{qwen3vl} & - & 83.5 & 87.6 & 42.8 & 47.3 & 60.8 & 72.2 & 65.7 \\
+ GUI-RCPO~\cite{du2026test} & SSv2 & 84.7 & 89.3 & 46.3 & 51.3 & 61.3 & 72.8 & 67.6\footnotesize{+1.9} \\
\rowcolor{gray!20} \textbf{+ Ours} & SSv2 & \textbf{88.5} & \textbf{91.6} & 47.6 & 52.1 & 61.2 & 75.4 & 69.4\footnotesize{+3.7} \\
+ GUI-RCPO~\cite{du2026test} & MMG & 84.2 & 86.9 & 42.3 & 46.8 & 61.8 & 71.6 & 65.6\footnotesize{-0.1} \\
\rowcolor{gray!20} \textbf{+ Ours} & MMG & 88.2 & 91.5 & \textbf{49.3} & \textbf{52.2} & \textbf{63.2} & \textbf{77.6} & \textbf{70.3\footnotesize{+4.6}} \\
\bottomrule
\end{tabular}%
\caption{GUI grounding accuracy on six benchmarks including ScreenSpot-V2 (SSv2), ScreenSpot-Pro (SSP), OSWorld-G (OSW-G), OSWorld-G\_R (OSW-GR), and MMBenchGUI (MMG). Bold indicates the best results.}
\label{tab:main}
\end{table*}

\subsection{Experimental Setup}
\noindent\textbf{Evaluation Benchmarks and Metrics.} We conduct experiments across diverse GUI visual grounding benchmarks covering various platforms and instruction complexities: ScreenSpot and ScreenSpot-v2, ScreenSpot-Pro, MMBench-GUI, as well as OSWorld-G and OSWorld-G-Refine. We adopt \emph{Element Accuracy} as the evaluation metric, where a prediction is correct if the predicted point falls within the bounding box of the target UI element.

\noindent\textbf{Implementation Details.} Our method is built upon the Qwen2.5-VL~\cite{qwen2.5-VL} and Qwen3-VL~\cite{qwen3vl}. To optimize memory efficiency, both the grounding model ($\pi_{\text{G}}$) and the Reflector ($\pi_{\text{R}}$) share a base model and are fine-tuned using LoRA~\cite{hu2022lora}. During training, we alternate between the roles of $\pi_{\text{G}}$ and $\pi_{\text{R}}$ by switching the active LoRA adapters. This parameter-efficient design restricts the GPU memory to approximately 10GB for 3B/2B models and 30GB for 7B/8B models.
During the rollout phase, we set the temperature to 1.0 and $\text{top-}p$ to 0.95. For the Reflector $\pi_{\text{R}}$, we train for 1 epoch via GRPO on 10K samples from GroundCUA~\cite{feizi2025grounding}, with a learning rate of $1 \times 10^{-4}$, a group size of 8, and a batch size of 64. For the internalization stage of $\pi_{\text{G}}$, R-OPSD integrates the GRPO advantage with a strength coefficient $\lambda = 0.2$. The model is trained for 2 epochs with a learning rate of $1 \times 10^{-4}$ and a batch size of 64. To simulate post-deployment self-evolving, we leverage data from ScreenSpot-v2 or MMBench-GUI, strictly omitting their ground-truth annotations.

\subsection{Main Results}
The evaluation results across six GUI visual grounding benchmarks are presented in Table~\ref{tab:main}. Overall, our proposed Test-Time Self-Evolving framework consistently enhances the grounding capabilities of the base models, outperforming the existing test-time adaptation method, \emph{e.g.}, GUI-RCPO.
Our method demonstrates performance gains across different base models. When deployed on the unseen SSv2 dataset, our framework boosts the average accuracy of Qwen2.5-VL-3B from 50.2\% to 57.4\%. When applied to the highly capable Qwen3-VL-2B backbone, our self-evolving framework still raises the average accuracy to 69.4\% (+3.7\%) and 70.3\% (+4.6\%) when adapting on SSv2 and MMG, respectively. This validates that our framework achieves the self-evolving without requiring ground-truth.

Adapting on complex datasets like MMG frequently results in rollout groups where all test-time explorations fail. GRPO used in GUI-RCPO relies on relative rewards, they fail to optimize from failed groups, causing negative transfer. In contrast, R-OPSD overcomes this limitation by leveraging reflections to provide token-level feedback, guiding the model even from complete failures. Supported by CC to filter out corrupted supervision, our framework achieves performance gains (+7.4\% and +4.6\%) with MMG.

\begin{table}[t!]
\centering
\small
\setlength{\tabcolsep}{1.5px}
\begin{tabular}{c|cccc|ccc|ccc}
\toprule
\multirow{2}{*}{} & \multirow{2}{*}{\textbf{R}} & \multirow{2}{*}{\textbf{CC}} & \multirow{2}{*}{\textbf{Clamp}} & \multirow{2}{*}{\textbf{QA}} & \multicolumn{3}{c|}{\textbf{Adapt on SSv2}} & \multicolumn{3}{c}{\textbf{Adapt on MMG}} \\
\cmidrule(lr){6-8} \cmidrule(lr){9-11}
& & & & & \textbf{SSv2} & \textbf{SSP} & \textbf{MMG} &\textbf{SSv2} & \textbf{SSP} & \textbf{MMG} \\ \midrule
 & \multicolumn{4}{c|}{\textbf{Base Model}} & 80.4 & 20.3 & 57.5 & 80.4 & 20.3 & 57.5 \\ \midrule
(a) & & & & &$\times$ & $\times$ & $\times$ & $\times$ & $\times$ & $\times$ \\
(b) &  &$\surd$ & & & 81.2 & 21.4 & 58.7 & 80.8 & 21.1 & 59.3 \\
(c) & & &$\surd$ & & 79.5 & 18.0 & 57.2 & 78.6 & 18.3 & 57.1 \\
(d) &  &$\surd$ & $\surd$ & & 84.2 & 24.6 & 61.7 & 82.3 & 23.5 & 61.5 \\
(e) &  &$\surd$ & $\surd$ & $\surd$ & 84.2 & 25.4 & 62.6 & 82.8 & 23.4 & 62.3 \\
(f) & $\surd$ & & & & $\times$& $\times$& $\times$& $\times$ & $\times$ & $\times$  \\
(g) & $\surd$ & $\surd$ & & & 84.0 & 26.8 & 62.9 & 84.0 & 27.7 & 64.3 \\
(h) & $\surd$ &  & $\surd$ & & 82.1 & 21.5 & 58.5 & 80.2 & 19.8 & 57.5 \\
(i) & $\surd$ & $\surd$ & $\surd$ & & 86.8 & 28.5 & 64.2 & 86.0 & 29.9 & 67.4 \\
(j) & $\surd$ & $\surd$ & $\surd$ & $\surd$ & \textbf{88.8} & \textbf{30.5} & \textbf{66.6} & \textbf{87.5} & \textbf{30.1} & \textbf{68.2} \\ \bottomrule
\end{tabular}
\caption{Ablation study on each component using Qwen2.5-VL-3B when adapting on SSv2 and MMG, respectively. \textbf{R}: Reflection $R$; \textbf{CC}: Contrastive Calibration Method; \textbf{Clamp}: Direction-based advantage clamping; \textbf{QA}: Integrating query-level advantage. $\times$ denotes suffering from policy collapse.}
\label{tab:ablation_main}
\end{table}

\begin{table}[t!]
\centering
\small
\setlength{\tabcolsep}{1.5px}
\begin{tabular}{c|c|ccc|ccc}
\toprule
\multirow{2}{*}{\textbf{Reward}} & \multirow{2}{*}{\textbf{Supervision}} & \multicolumn{3}{c|}{\textbf{Adapt on SSv2}} & \multicolumn{3}{c}{\textbf{Adapt on MMG}} \\
\cmidrule(lr){3-5} \cmidrule(lr){6-8}
& & \textbf{SSv2} & \textbf{SSP} & \textbf{MMG} &\textbf{SSv2} & \textbf{SSP} & \textbf{MMG} \\ \midrule
 \multicolumn{2}{c|}{\textbf{Base Model}} & 80.4 & 20.3 & 57.5 & 80.4 & 20.3 & 57.5 \\ \midrule
GUI-RCPO & - & 85.4 & 24.8 & 60.2 & 80.6 & 19.8 & 58.2 \\
Binary & \emph{Avg.} $B$ & 84.3 & 24.6 & 61.7 & 79.1 & 21.2 & 57.5 \\
IoU & \emph{Avg.} $B$ & 84.7 & 22.4 & 61.3 & 80.8 & 20.4 & 59.4 \\
GUI-G$^{2}$ & \emph{Avg.} $B$ & 83.0 & 21.2 & 59.9 & 78.6 & 19.4 & 55.8 \\
Binary & $S$ & 85.4 & 25.9 & 62.3 & 84.2 & 25.7 & 62.8 \\ \midrule
GUI-SD & \emph{Avg.} $B$ & 78.5 & 16.7 & 52.3 & $\times$ & $\times$ &$\times$ \\ \midrule
\rowcolor{gray!20} \textbf{Ours} & $S,R$ & \textbf{88.8} & \textbf{30.5} & \textbf{66.6} & \textbf{87.5} & \textbf{30.1} & \textbf{68.2} \\ \bottomrule
\end{tabular}
\caption{Comparison of the performance between GRPO with various scalar rewards and R-OPSD. The ``Supervision'' denotes the source of feedback used for optimization: \emph{Avg.} $B$ indicates that the rewards are derived from aggregated bounding box predictions within a rollout group; $S$ is the evaluation results from the Reflector.}
\label{tab:ablation_opd}
\end{table}

\begin{figure}
\centering
\begin{tikzpicture}
    \begin{groupplot}[
        group style={
            group size=2 by 1,          
            horizontal sep=0.1\linewidth, 
            vertical sep=0pt,
            xlabels at=edge bottom,     
        },
        width=0.55\linewidth,           
        height=3.0cm,                   
        xmin=0, xmax=0.5,
        xtick={0, 0.1, 0.2, 0.3, 0.4, 0.5},
        tick label style={font=\scriptsize},
        label style={font=\small},
        title style={font=\small, yshift=-1.5ex}, 
        grid=major,
        grid style={dashed, gray!30},
        legend style={
            font=\tiny,
            fill=white,
            draw=black!50,
            inner sep=2pt,
            at={(0.78,0.02)},          
            anchor=south east,
        },
    ]

    \nextgroupplot[
        ylabel={Accuracy (\%)},         
        title={SSv2},
        ymin=80, ymax=90,               
        ytick={80, 82, 84, 86, 88, 90},
    ]
    \addplot[
        color=blue,
        mark=square*,
        mark size=1.8pt,
        line width=0.8pt
    ]
    coordinates {(0.0, 85.4) (0.1, 86.7) (0.2, 88.8) (0.3, 87.9) (0.4, 85.7) (0.5, 85.8)};
    \addlegendentry{SSv2}
    \addplot[
        color=red,
        mark=triangle*,
        mark size=2pt,
        line width=0.8pt
    ]
    coordinates {(0.0, 84.2) (0.1, 85.2) (0.2, 87.5) (0.3, 87.4) (0.4, 84.8) (0.5, 84.4)};
    \addlegendentry{MMG}

    \nextgroupplot[
        ymin=58, ymax=70,               
        title={MMG},
        ytick={58, 62, 66, 70},
    ]
    \addplot[
        color=blue,
        mark=square*,
        mark size=1.8pt,
        line width=0.8pt
    ]
    coordinates {(0.0, 62.3) (0.1, 63.7) (0.2, 66.6) (0.3, 65.4) (0.4, 65.7) (0.5, 63.6)};
    \addlegendentry{SSv2}
    \addplot[
        color=red,
        mark=triangle*,
        mark size=2pt,
        line width=0.8pt
    ]
    coordinates {(0.0, 62.8) (0.1, 64.9) (0.2, 68.2) (0.3, 67.0) (0.4, 64.5) (0.5, 63.3)};
    \addlegendentry{MMG}

    \end{groupplot}
\end{tikzpicture}
\caption{Sensitivity analysis of the integration strength $\lambda$. The subplots indicate the evaluation datasets. The colored lines denote the adaptation datasets.}
\label{fig:lambda_ablation}
\end{figure}
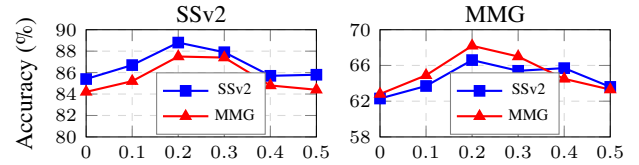

\subsection{Ablation Study}
Qualitative and quantitative analysis of the Reflector, computational overhead analysis, scalability to larger base models, and comparisons with visual bounding box prompting can be found in the \textbf{Appendix}.

\noindent\textbf{Effectiveness of Each Component.}
To validate the effectiveness of each module in our Test-Time Self-Evolving framework, we conduct ablation studies in Table~\ref{tab:ablation_main}.
We establish (a) as our baseline, which applies OPSD using only evaluation results as privileged information. This naive baseline suffers from catastrophic policy collapse.
Comparing (d)(e) and (i)(j), incorporating Reflection yields a substantial performance gain, \emph{e.g.}, when adapting on SSv2, Reflection boosts the accuracy from 24.6\% to 28.5\% on SSP.
Utilizing token-level feedback in auto-regressive generation introduces the incorrect prefix issue. (f) utilizes reflection but lacks CC, causing it to collapse. By incorporating CC, (g) filters out this corrupted supervision, turning the negative transfer into massive gains, reaching 64.3\% on MMG. A similar substantial boost is observed in (h) \emph{vs.} (i), confirming CC is important for learning from failed explorations.
(i) demonstrates that direction-based advantage clamping prevents token-level updates from contradicting the evaluation result, providing consistent improvements. Building upon this, (j) shows the benefit of integrating the query-level advantage.

\noindent\textbf{Necessity of R-OPSD over Standard RL.}
To demonstrate the necessity of R-OPSD, we compare it against GRPO using various scalar rewards in Table~\ref{tab:ablation_opd}. Following TTRL~\cite{zuo2026ttrl}, relying on averaged predicted bounding boxes (\emph{Avg} $B$) as pseudo-labels suffers negative transfer on complex datasets like MMG. This occurs because when some predictions drift towards visually similar but incorrect UI elements, averaging these bounding boxes leads to completely invalid pseudo-labels. The evaluation result $S$ from Reflector is more robust to this situation, leading to better performance. This also confirms the high reliability of the Reflector. However, GRPO guided by the binary $S$ reward remains limited by the sparsity of scalar signals and cannot instruct the model on how to correct specific coordinate tokens.

Furthermore, to verify that our gains stem from the reflection design rather than OPSD, we evaluate GUI-SD~\cite{zhang2026learn}, which draws the bounding box in the image as the privileged information. Its severe performance degradation confirms that drawing invalid pseudo-labels directly corrupts the visual context.
In contrast, R-OPSD leverages both the evaluation result and the detailed textual reflection ($S, R$), achieving the highest accuracy.

\noindent\textbf{Sensitivity Analysis of Integration Strength.}
We investigate the impact of $\lambda$, which controls the magnitude of the distillation advantage during policy updates. As shown in Fig.~\ref{fig:lambda_ablation}, $\lambda = 0.2$ leads to the best performance, regardless of the adaptation and evaluation datasets.

\section{Conclusion}
In this paper, we introduce a Test-Time Self-Evolving framework for GUI visual grounding that enables post-deployment adaptation without requiring ground truth annotations. Our framework introduces a closed-loop of Exploration, Evaluation, Reflection, and Internalization, allowing the model to learn from both the outcomes of its explorations and the reasons behind its successes and failures. To internalize such reflection knowledge, we propose R-OPSD, which transforms natural-language reflections into token-level supervision through a self-teacher conditioned on privileged information. Furthermore, we introduce a Contrastive Calibration method to alleviate corrupted supervision caused by incorrect auto-regressive prefixes during failed explorations. Extensive experiments across diverse benchmarks and ablations demonstrate that our framework can improve the performance of the model in a self-evolving manner.

\bibliography{aaai2027}

\clearpage
\section*{Appendix}
This Appendix provides the details of construction of the Reflector training dataset, additional qualitative and quantitative analysis of the Reflector, computational overhead analysis, scalability to larger base models, extension of our method to the unsupervised training setting, and comparisons with visual bounding box prompting.

\subsection{Construction of the Reflector Training Dataset}
\label{sec:appendix_reflector_data}
To train the MLLM-based Reflector $\pi_{\text{R}}$ to accurately evaluate test-time explorations and provide diagnostic reasoning, we construct a training dataset based on the \emph{Functional Split} of GroundCUA~\cite{feizi2025grounding}. \textbf{The constructed Reflector training dataset will be released.} It is designed to capture a diverse set of predictions while maintaining balanced labels, proceeding as follows:

\noindent\textbf{Prediction Generation.} For each instruction-image pair in the GroundCUA, we employ the base model to perform inference. To encourage a diverse range of predictions, we set the sampling temperature to $1.0$ and sample 8 distinct coordinate predictions per instruction.

\noindent\textbf{Label Assignment.} Each generated coordinate is evaluated against the ground-truth bounding box. A discrete binary label $S^* \in \{0, 1\}$ is assigned to each prediction, where $S^* = 1$ indicates a successful grounding (the predicted point falls within the target bounding box) and $S^* = 0$ indicates a failure.

\noindent\textbf{Sample Filtering.} We group the 8 predictions by their instruction and apply a filtering strategy: we exclusively retain the groups that exhibit mixed results. Specifically, an instruction is kept only if its 8 rollouts contain at least one correct prediction $S^*=1$ and at least one incorrect prediction $S^*=0$.

\noindent\textbf{Balanced Sampling.} To ensure a balanced distribution of positive (``Yes'') and negative (``No'') evaluation results, we randomly select exactly one correct prediction $S^*=1$ and one incorrect prediction $S^*=0$ from the 8 generated rollouts for each retained instruction.

The final Reflector training dataset consists of approximately $10,160$ prediction-label pairs, yielding a 1:1 ratio of positive to negative samples.

\noindent\textbf{Discussion on Cold-Start SFT.} Additionally, we have explored a supervised fine-tuning cold-start strategy by synthesizing 30K high-quality reasoning trajectories using Qwen3.7-plus. However, results indicated that incorporating this cold-start phase yielded no noticeable performance gain compared to directly applying RL to the base model for the training of the Reflector.

\subsection{More Experimental Results}
\noindent\textbf{Reliability Evaluation of the MLLM-based Reflector.}
The efficacy of our Test-Time Self-Evolution framework relies heavily on the quality of the evaluation score $S$ and diagnostic reasoning $R$ provided by the Reflector $\pi_{\text{R}}$. We evaluate the Reflector on a held-out test set consisting of 1,000 randomly sampled instruction-prediction pairs from SSv2, SSP, and MMG. We maintain a balanced 1:1 ratio of successful and failed groundings. We assess its performance as a binary classifier $S \in \{0, 1\}$ using three metrics: Accuracy, Precision, and Recall. Here, the positive class represents successful groundings $S^*=1$.

\begin{table}[h]
\centering
\small
\setlength{\tabcolsep}{3pt}
\begin{tabular}{l c c c}
\toprule
\textbf{Model} & \textbf{Accuracy} & \textbf{Precision} & \textbf{Recall} \\
\midrule
Qwen2.5-VL-3B (Zero-shot) & 76.5 & 72.9 & 85.4 \\
\rowcolor{gray!20} \textbf{Trained Reflector} $\pi_{\text{R}}$ & \textbf{89.5} & \textbf{86.9} & \textbf{93.3} \\
Qwen3-VL-2B (Zero-shot) & 80.5 & 78.9 & 86.0 \\
\rowcolor{gray!20} \textbf{Trained Reflector} $\pi_{\text{R}}$ & \textbf{91.7} & \textbf{88.8} & \textbf{95.7} \\
\bottomrule
\end{tabular}
\caption{Performance metrics of the MLLM-based Reflector.}
\label{tab:reflector_performance}
\end{table}

As shown in Table~\ref{tab:reflector_performance}, after training via GRPO with format and binary rewards, our Reflector achieves an evaluation Accuracy of 89.5\% and 91.7\%, demonstrating its capability in determining whether a predicted coordinate aligns with the user instruction.

\noindent\textbf{Computational Overhead and Training Time Analysis.} To provide a comprehensive understanding of the computational overhead, we conduct a training time analysis using the Qwen2.5-VL-3B model with 4 NVIDIA A100 40GB GPUs.
As shown in Table~\ref{tab:time_ablation}, to compute relative advantages, GRPO samples a group of trajectories $K=8$ for every single query, leading to the total training time of 449 minutes.
R-OPSD operates on a different mechanism. By utilizing the Reflector's feedback as privileged information to construct a self-teacher, R-OPSD performs a token-level knowledge distillation. The model only needs to sample a single trajectory per query. Even with the additional forward passes required for the self-teacher and the inverse-prompted student in Contrastive Calibration (CC), the R-OPSD w/ CC only takes 151 minutes, reducing the training time by more than 34\% compared to GRPO.

Our full framework (R-OPSD w/ CC \& QA) achieves the optimal performance by integrating the token-level advantage. Re-introducing QA necessitates returning to the $K=8$ group rollout setting, which increases the training time to 778 minutes ($1.73\times$).
For resource-constrained deployment environments, R-OPSD w/ CC serves as a highly efficient learning method that operates at $1 / 3$ the cost of standard RL. The full framework leverages the combined strengths of both token-level and query-level advantages, offering a computational trade-off.

\begin{table}
\centering
\small
\setlength{\tabcolsep}{6pt}
\begin{tabular}{l | c c}
\toprule
\textbf{Method} & \textbf{Time (m)} & \textbf{Relative Cost} \\
\midrule
GRPO + GUI-RCPO & 449 & $1.00\times$ \\
GRPO + Binary & 447 & $1.00\times$ \\ \midrule
R-OPSD (w/o CC) & 129 & $\approx 0.29\times$ \\
R-OPSD (w/ CC) & 151 & $\approx 0.34\times$ \\
R-OPSD (w/ CC \& QA) & 778 & $\approx 1.73\times$ \\
\bottomrule
\end{tabular}
\caption{Training time Analysis. The Relative Cost is normalized against the standard GRPO baseline.}
\label{tab:time_ablation}
\end{table}

\noindent\textbf{Scalability to Larger Base Models.}
In the main text, our experiments are conducted on the 3B and 2B parameter scales, \emph{e.g.,} Qwen2.5-VL-3B and Qwen3-VL-2B. To further demonstrate the scalability of our Test-Time Self-Evolution framework, we extend our evaluation to a larger base model. The evaluation results are shown in Table~\ref{tab:scaling_7b}. The Qwen2.5-VL-7B and Qwen3-VL-8B base models exhibit stronger initial grounding capabilities than 3B/2B versions. However, even with this baseline, our framework consistently yields performance improvements. For instance, adapting the 7B model on MMG using our framework boosts the accuracy from 68.2\% to 79.2\%. Our method still outperforms the TTRL method GUI-RCPO. These results further demonstrate the scalability of our framework.

\begin{table}
\centering
\small
\setlength{\tabcolsep}{4pt}
\begin{tabular}{l | ccc | ccc}
\toprule
\multirow{2}{*}{\textbf{Method}} & \multicolumn{3}{c|}{\textbf{Adapt on SSv2}} & \multicolumn{3}{c}{\textbf{Adapt on MMG}} \\
\cmidrule(lr){2-4} \cmidrule(lr){5-7}
& \textbf{SSv2} & \textbf{SSP} & \textbf{MMG} & \textbf{SSv2} & \textbf{SSP} & \textbf{MMG} \\
\midrule
Qwen2.5-VL-7B & 86.8 & 19.9 & 68.2 & 86.8 & 19.9 & 68.2 \\
+ GUI-RCPO & 89.2 & 25.9 & 70.3 & 88.3 & 25.5 & 70.9 \\
\rowcolor{gray!20} \textbf{+ Ours} & \textbf{92.7} & \textbf{31.4} & \textbf{77.9} & \textbf{92.5} & \textbf{32.3} & \textbf{79.2} \\ \midrule
Qwen3-VL-8B & 92.9 & 53.5 & 82.7 & 92.9 & 53.5 & 82.7 \\
+ GUI-RCPO & 93.2 & 53.8 & 83.3 & 93.5 & 54.6 & 83.4 \\
\rowcolor{gray!20} \textbf{+ Ours} & \textbf{94.5} & \textbf{55.8} & \textbf{84.3} & \textbf{94.1} & \textbf{56.3} & \textbf{84.7} \\
\bottomrule
\end{tabular}
\caption{Scalability evaluation using Qwen2.5-VL-7B and Qwen3-VL-8B base models. We report the target dataset accuracy when adapting on SSv2 and MMG, respectively.}
\label{tab:scaling_7b}
\end{table}

\noindent\textbf{Comparison with Visual Bounding Box Prompting.}
Many methods directly draw a bounding box or point on the UI screenshot for representing grounding coordinates~\cite{zhang2026learn,luo2025visual}. To evaluate this design, we conduct an ablation study comparing textual coordinate representation against visual bounding box representation in two critical components of our framework: the Reflector's evaluation and the teacher model's privileged information prompt.
For the Reflector + V and Ours + V, instead of feeding the textual coordinate token $B$ in the prompt, we draw a red bounding box corresponding to the prediction on the input image $I$.

\begin{table}
\centering
\small
\setlength{\tabcolsep}{3pt}
\begin{tabular}{l c c c}
\toprule
\textbf{Model} & \textbf{Accuracy} & \textbf{Precision} & \textbf{Recall} \\
\midrule
Qwen2.5-VL-3B (Zero-shot) & 76.5 & 72.9 & 85.4 \\
\rowcolor{gray!20} \textbf{Trained Reflector} $\pi_{\text{R}}$ & \textbf{91.7} & \textbf{88.8} & \textbf{95.7} \\
Qwen2.5-VL-3B + V (Zero-shot) & 73.2 & 73.5 & 78.6 \\
\rowcolor{gray!20} \textbf{Trained Reflector + V} $\pi_{\text{R}}$ & \textbf{78.3} & \textbf{78.5} & \textbf{79.5} \\
\bottomrule
\end{tabular}
\caption{Performance comparison of the Reflector with visual bounding box prompt and textual coordinate prompt.}
\label{tab:reflector_comparison}
\end{table}

\begin{table}
\centering
\small
\setlength{\tabcolsep}{4pt}
\begin{tabular}{l | ccc | ccc}
\toprule
\multirow{2}{*}{\textbf{Method}} & \multicolumn{3}{c|}{\textbf{Adapt on SSv2}} & \multicolumn{3}{c}{\textbf{Adapt on MMG}} \\
\cmidrule(lr){2-4} \cmidrule(lr){5-7}
& \textbf{SSv2} & \textbf{SSP} & \textbf{MMG} & \textbf{SSv2} & \textbf{SSP} & \textbf{MMG} \\
\midrule
Qwen2.5-VL-3B & 80.4 & 20.3 & 57.5 & 80.4 & 20.3 & 57.5 \\
\rowcolor{gray!20} \textbf{+ Ours} & 88.8 & 30.5 & 66.6 & 87.5 & 30.1 & 68.2 \\
\rowcolor{gray!20} \textbf{+ Ours + V} & 77.3 & 18.3 & 53.2 & 77.5 & 17.4 & 54.7 \\
\bottomrule
\end{tabular}
\caption{Performance comparison of our method with visual bounding box prompt and textual coordinate prompt.}
\label{tab:visual_vs_text}
\end{table}

Table~\ref{tab:reflector_comparison} summarizes the evaluation results. Our trained Reflector with textual coordinates achieves a remarkable accuracy of 91.7\%, outperforming the zero-shot base model by 15.2\%. Conversely, employing visual prompts (+ V), the Trained Reflector + V only reaches 78.3\% accuracy. The negative impact of visual prompting is amplified during the parameter internalization stage. As demonstrated in Table~\ref{tab:visual_vs_text}, leveraging drawn images (Ours + V) as the privileged visual context for the self-teacher in R-OPSD leads to negative transfer.

\noindent\textbf{Extension beyond Test-Time Data.}
While our Test-Time Self-Evolving framework is designed to adapt models to unseen interfaces after deployment, its core internalization mechanism, R-OPSD, inherently serves as an unsupervised training algorithm.
Instead of episodic exploration on test sets, we train the base model directly on the training set of GUI-R1~\cite{gui-r1}, which contains about 3000 samples. To simulate the unsupervised training setting, we only leverage the image and instruction. The results are summarized in Table~\ref{tab:gui-r1}. It is shown that our method consistently outperforms the base model across all three datasets, demonstrating its effectiveness in unsupervised training setting.

\begin{table}
\centering
\small
\setlength{\tabcolsep}{4pt}
\begin{tabular}{l | ccc}
\toprule
\multirow{2}{*}{\textbf{Method}} & \multicolumn{3}{c}{\textbf{GUI-R1 Training Set}} \\
\cmidrule(lr){2-4}
& \textbf{SSv2} & \textbf{SSP} & \textbf{MMG} \\
\midrule
Qwen2.5-VL-3B & 80.4 & 20.3 & 57.5 \\
\rowcolor{gray!20} \textbf{+ Ours} & \textbf{87.8} & \textbf{29.2} & \textbf{67.2} \\
\bottomrule
\end{tabular}
\caption{Performance of our method on the unsupervised training setting.}
\label{tab:gui-r1}
\end{table}

\subsection{Qualitative Analysis of the Reflector}
\label{sec:appendix_reflector_vis}
\begin{figure*}[t!]
    \centering
    \begin{subfigure}{0.49\textwidth}
        \centering
        \includegraphics[width=\linewidth]{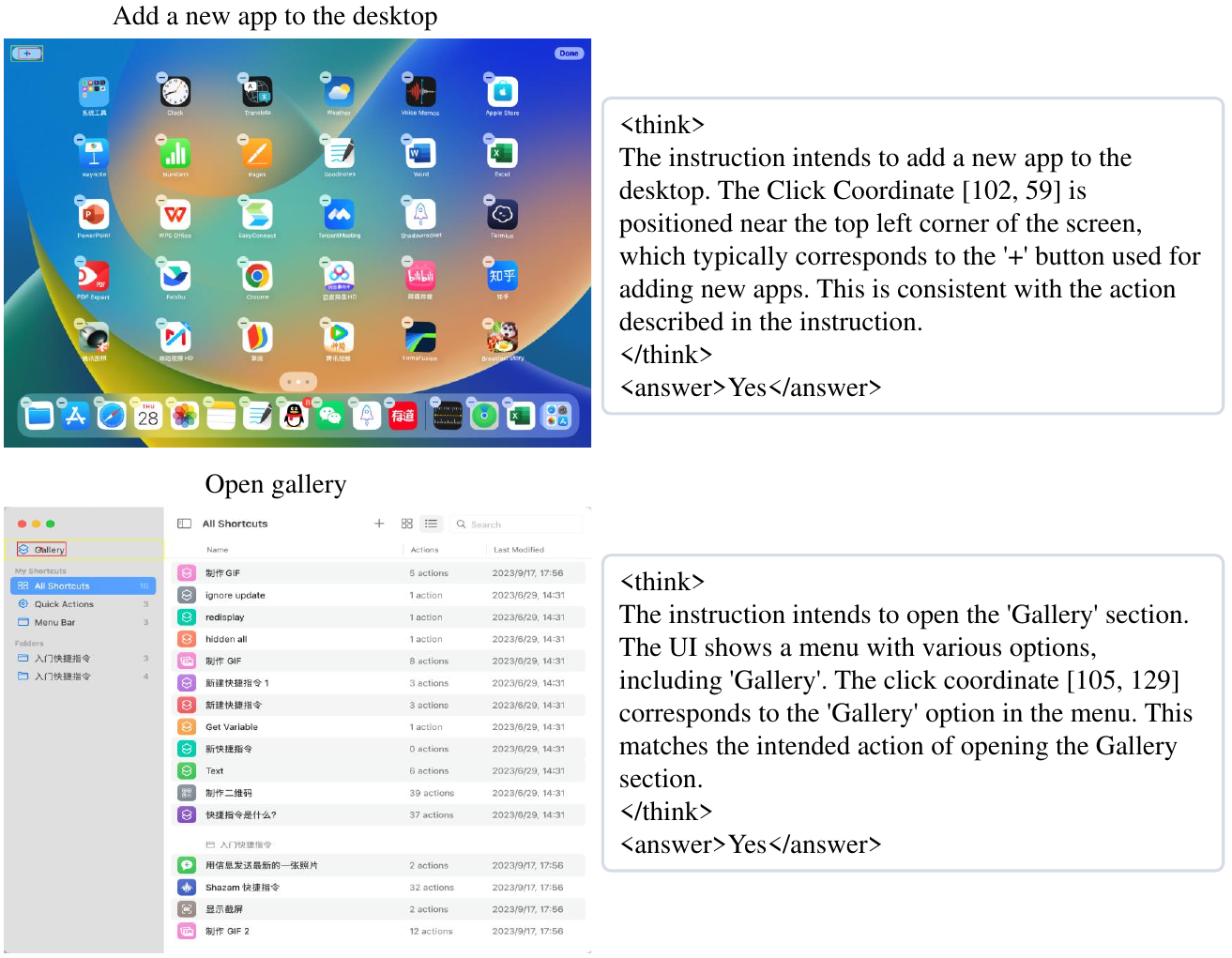}
        \caption{Successful Exploration ($S=1$)}
        \label{fig:reflector_success}
    \end{subfigure}
    \begin{subfigure}{0.49\textwidth}
        \centering
        \includegraphics[width=\linewidth]{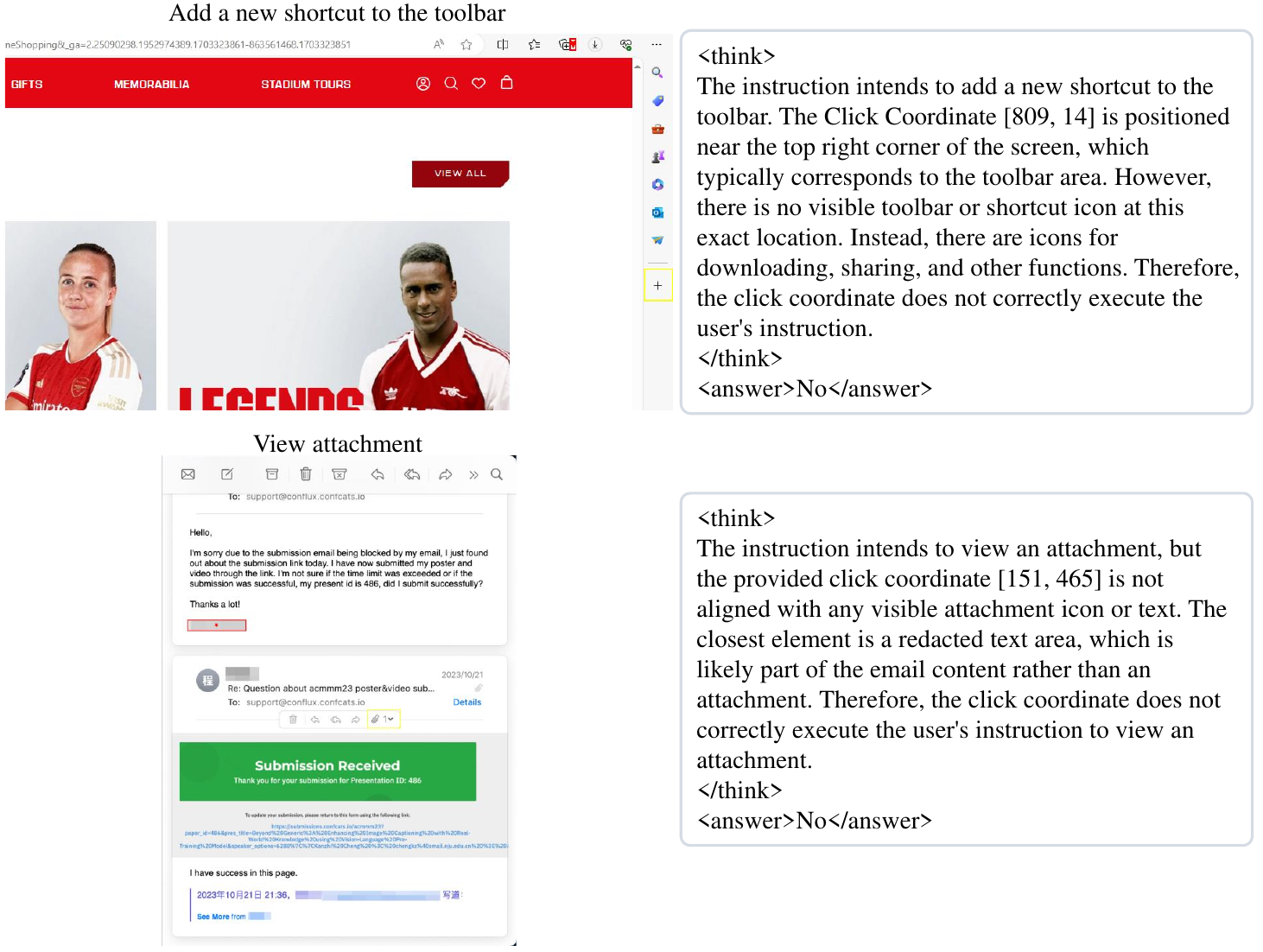}
        \caption{Failed Exploration ($S=0$)}
        \label{fig:reflector_failure}
    \end{subfigure}
    \caption{Qualitative visualizations of the Reflector's evaluation and reasoning. The red bounding boxes denote the predicted coordinates. The yellow bounding boxes highlight the corresponding UI element.}
    \label{fig:reflector_vis}
\end{figure*}

To provide an intuitive understanding of the Reflector's capabilities, we present qualitative visualizations of its evaluation process in Fig.~\ref{fig:reflector_vis}. The visualization showcases the Reflector's multi-modal reasoning across two typical test-time exploration scenarios: a successful grounding ($S=1$) and a failed grounding ($S=0$).
As illustrated in the successful case, the Reflector accurately parses the user's intent, spatially grounds the agent's predicted coordinate to the corresponding UI element, and logically confirms the match. More importantly, in the failed case, the Reflector does not merely output a binary rejection. Instead, it explicitly pinpoints the exact cause of the failure within its step-by-step reflection $R$. 

\end{document}